\documentclass[runningheads]{llncs}
\usepackage[table]{xcolor} 
\usepackage{makecell}
\usepackage[T1]{fontenc}
\usepackage{graphicx}
\usepackage{amsmath}
\usepackage{etoolbox}
\makeatletter
\renewcommand*\inst[1]{\textsuperscript{#1}}
\makeatother

\newcommand{\corrauth}{\textsuperscript{*}}
\begin{document}
\title{Construction of Knowledge Graph based on Language Model}
%
%
\author{Quibai Zhu\inst{1,2}\orcidID{0009-0004-2587-5310} \and
Qingwang Wang\inst{1,2}\corrauth\orcidID{0000-    0001-5820-5357} \and
Haibin Yuan\inst{3} \and
Wei Chen\inst{1}\orcidID{0000-0002-3169-782X} \and
Tao Shen\inst{1,2}\orcidID{0000-0003-1273-7950}}
\authorrunning{Q. Zhu et al.}
%
\institute{
  \textsuperscript{1} Faculty of Information Engineering and Automation, Kunming University of Science and Technology, Kunming 650500, China \\
  \textsuperscript{2} Yunnan Key Laboratory of Computer Technologies Application, Kunming University of Science and Technology, Kunming 650500, China \\
  \textsuperscript{3} Tin Branch, Yunnan Tin Co., LTD.\\
  \email{wangqingwang@kust.edu.cn}
}
\maketitle              
\begin{abstract}
Knowledge Graph (KG) can effectively integrate valuable information from massive data, and thus has been rapidly developed and widely used in many fields. Traditional KG construction methods rely on manual annotation, which often consumes a lot of time and manpower. And KG construction schemes based on deep learning tend to have weak generalization capabilities. With the rapid development of Pre-trained Language Models (PLM), PLM has shown great potential in the field of KG construction. This paper provides a comprehensive review of recent research advances in the field of construction of KGs using PLM. In this paper, we explain how PLM can utilize its language understanding and generation capabilities to automatically extract key information for KGs, such as entities and relations, from textual data. In addition, We also propose a new Hyper-Relarional Knowledge Graph construction framework based on lightweight Large Language Model (LLM) named LLHKG and compares it with previous methods. Under our framework, the KG construction capability of lightweight LLM is comparable to GPT3.5.
\end{abstract}
\section{Introduction}
Knowledge Graph (KG) was first proposed by Google in 2012 \cite{ref1}. KG is a structured semantic network for representing and storing knowledge. Its organizes knowledge through a graph structure of entities, relations and attributes. Among them, the entity is the basic unit of KG, the relation describes the semantic connection between entities, and the attribute is used to describe the characteristics of the entity. KG has been widely applied and achieved remarkable results in numerous vertical fields, including intelligent search \cite{ref2}, recommender systems \cite{ref3} and financial risk assessment \cite{ref4}.

The knowledge graph construction scheme based on deep learning mainly automatically extracts entities and relations from a large amount of text 
\begin{figure}
\includegraphics[width=\textwidth]{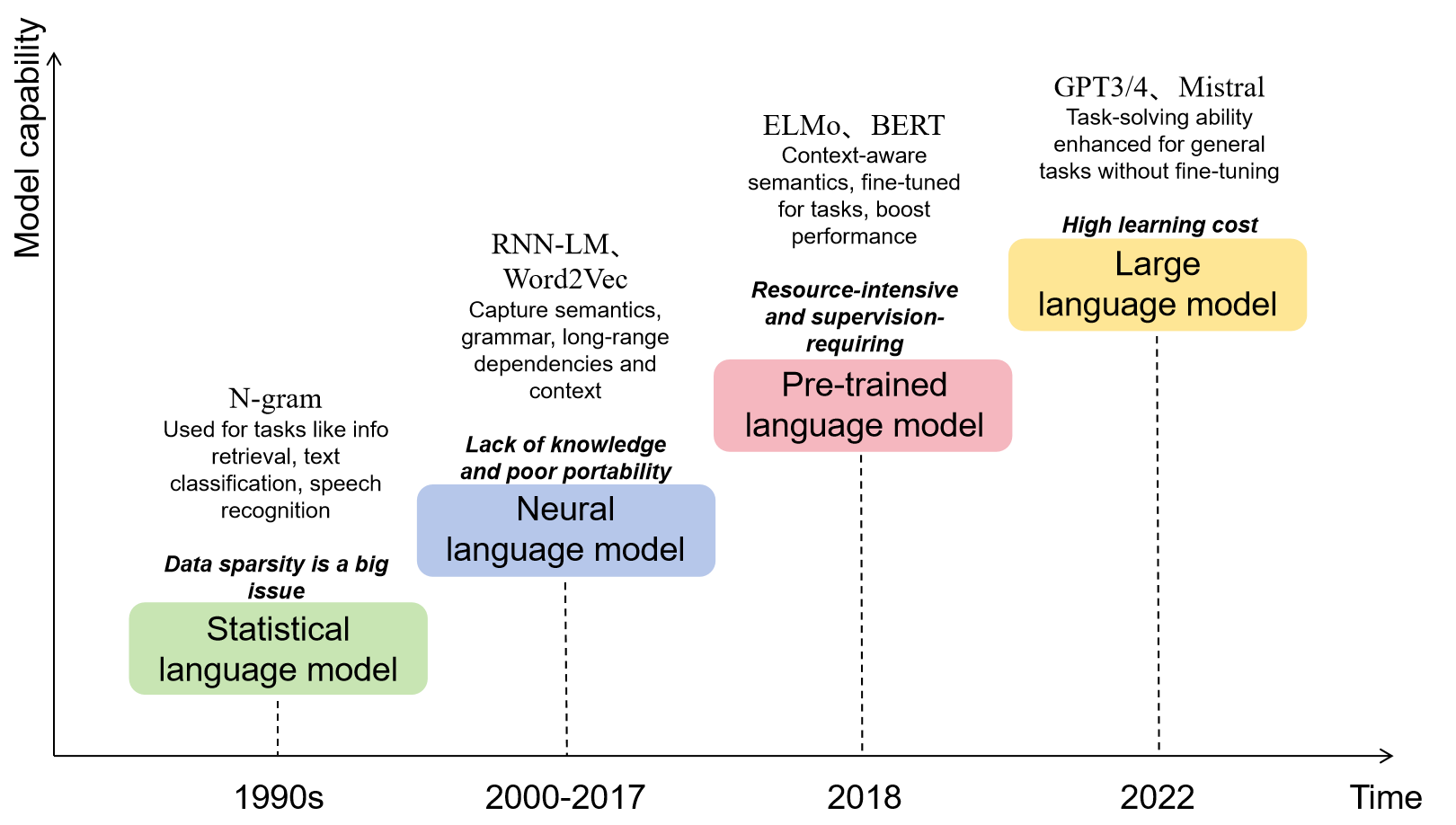}
\caption{The development of language models The development process of language models, representative models, model capabilities and existing shortcomings are briefly described. } \label{fig1}
\end{figure}
through neural network models. These schemes usually include steps such as text preprocessing, named entity recognition, relation extraction, knowledge fusion, and knowledge storage and update. They are capable of handling large-scale data and automatically learning complex patterns and relationships through deep learning models. However, these methods usually require a large amount of labeled data for training, demand high data quality and labeling quality. Moreover, these methods face challenges when dealing with complex multi-hop relationships and long texts.

As shown in Fig. \ref{fig1}, language models have evolved from early statistical methods to modern deep learning techniques. Early models relied on basic statistics, but with neural networks like Word2Vec, they began capturing language features. Recently, Pre-trained Language Model (PLM) such as ELMo and BERT have significantly improved understanding and generation via pre-training on large texts. The latest large language model (LLM) has further enhanced task-solving abilities, reduced task-specific data dependence, and enabled strong performance in various NLP tasks.

In recent years, PLM has made remarkable progress in the field of NLP. Its powerful language generation and understanding capabilities provide new ideas for the construction and application of KG. Compared with the knowledge graph construction scheme based on deep learning, the scheme based on PLM has significant advantages. PLM is pre-trained on a large amount of unlabeled text and has low requirements for labeled data, which makes it suitable for data-scarce scenarios. In this paper, we focus on analyzing PLM-based KG construction schemes and propose a novel HRKG construction framework called LLHKG. In our framework, the KG construction capability of lightweight LLM is comparable to GPT3.5.

\section{KG Construction based on PLM }
\subsection{KG Construction Scheme based on Language Model}
The current KG construction schemes based on language models can be divided into two main categories: mining the internal knowledge of language models and KG extraction based on PLMs. Each of these approaches focuses on the depth and breadth of knowledge extraction, providing diversified technical paths for the construction of KGs.

\subsubsection{Mining the Internal Knowledge of Language Models} This type of scheme focuses on exploring the internal knowledge structure of language models. By analyzing the internal mechanisms of language models, researchers are able to identify and extract information such as entities, relations and attributes in the KG. For example, Wang et al. \cite{ref5} regarde language models as open KGs, emphasizing that a large amount of knowledge has been implied inside them, which can be mined and exploited through appropriate methods. Cohen et al. \cite{ref6} further explore the implicit knowledge structure of language models by crawling their internal knowledge base. The core of such methods is to deeply understand the internal representation of language models in order to realize knowledge extraction.

\subsubsection{KG Extraction based on PLMs} Different from mining internal knowledge, this type of scheme focuses on utilizing the output or internal representation of PLMs to generate KGs through specific techniques or algorithms. For example, Hao et al. \cite{ref7} utilize the internal representation of the pre-trained model BERT and transform this knowledge into explicit KG structures through specific algorithms. Wang et al. \cite{ref8} treat the information extraction task as a text-to-triad translation problem, and extract the structured information of the KG directly from the text through the generative capability of the PLM. These methods have the advantage of being able to quickly construct KG by directly utilizing the powerful generative capabilities of pre-trained models.

\subsection{Knowledge mining based on LLM}
Current knowledge mining based on LLMs can be categorized into prompt-based knowledge mining, zero-shot or multi-example knowledge mining, and domain-specific knowledge mining.

\subsubsection{Prompt-based Knowledge Mining} Related studies guide LLMs to knowledge mining by designing and optimizing hints. For example, Gan et al. \cite{ref9} optimize the knowledge mining ability of LLMs for online marketing by using stepwise enhancement prompt techniques. This scheme significantly improves the model performance and reduces the deployment cost. Wu et al. \cite{ref10} compares In-Context Learning (ICL) and Prompting Based Factual Knowledge Extraction (PFK) and finds that PFK is more accurate in some tasks but requires higher quality of prompts, while ICL relies less on prompts but may not be accurate enough in complex tasks.

\subsubsection{Knowledge Mining with Zero Prompt or Multiple Examples} Related studies work on reducing the reliance on prompt. Wu et al. \cite{ref11} propose a zero-shot multiple-example approach to simplify the knowledge mining process by guiding the model to extract knowledge through multiple examples. In addition, Kang et al. \cite{ref12} conduct an empirical study on zero-shot keyphrase extraction, and found that Llama3 and Gemma2 perform well in the zero-shot keyphrase extraction task, with mixed prompts consistently improving results in most LLMs.

\subsubsection{Domain-specific Knowledge Mining} In specific domains such as scientific text, Dagdelen et al. \cite{ref13} propose a joint named entity recognition and relationship extraction methodology. It shows how LLMs can be fine-tuned to extract complex scientific knowledge, providing a simple and flexible way to extract structured specialized scientific knowledge from research papers.

\subsection{KG Construction based on LLM }
Current research on LLM-based KG construction has advanced significantly in zero-shot and few-shot learning, extracting knowledge from large texts to generate structured graphs. Meanwhile, in terms of domain-specific knowledge graph construction, researchers have designed efficient construction methods for different domain data characteristics. Additionally, iterative validation techniques have improved knowledge graph quality and reliability through innovative mechanisms and optimizations.

\subsubsection{Zero-shot Learning and Few-shot Learning} Researchers are leveraging LLMs to construct high-quality knowledge graphs without extensive labeled data. Zhang et al. \cite{ref14} uses LLMs to generate knowledge graphs through three phases: information extraction, schema definition, and normalization, without needing predefined schemas. This method is adaptable to various domains and enhances construction efficiency and accuracy. Additionally, Carta et al. \cite{ref15} employs an iterative zero-shot prompting strategy with LLMs (e.g., GPT-3.5) to extract knowledge graph components, requiring no external resources and offering high scalability.

\subsubsection{Domain-Specific Knowledge Graph Construction} Researchers are leveraging the generative capabilities of LLMs to develop efficient construction methods tailored to the characteristics of domain-specific data. For instance, Bai et al. \cite{ref16} utilizes LLM-generated data in conjunction with the Neo4j database, employing a Retrieval-Augmented Generation (RAG) process to achieve efficient construction and application of knowledge graphs, thereby significantly enhancing the organization and utilization efficiency of domain-specific knowledge. Additionally, Chen et al. \cite{ref17} introduce the SAC-KG framework, which comprises three components—Generator, Verifier, and Pruner—to automate the construction of large-scale, high-precision multilevel knowledge graphs from domain corpora, thereby minimizing human intervention and enhancing practical applicability.

\subsubsection{KG Construction based on Iterative Validation} PiVe \cite{ref18} enhances LLMs' KG construction via iterative verification, ensuring accurate and reliable content while reducing errors and illusions. In addition to zero-sample learning, Carta et al. \cite{ref15} also emphasizes the importance of iterative validation. High-quality KGs are constructed through multiple iterations and validation to ensure that the knowledge components extracted each time are accurate.

\section{Construction of HRKG and Knowledge Hypergraph based on LLM}
Traditional KGs represent knowledge through triples, but they struggle with many-to-many relations. Hyper-Relational Knowledge Graph (HRKG) addresses this by using hyperedges to connect multiple nodes simultaneously, more naturally representing complex multi-body relations.

There are few papers on constructing HRKG using LLM. In 2024, Datta et al. \cite{ref19} proposed a scheme using LLM to construct HRKG via prompt engineering and CoT with GPT3.5, which is referred to as GCLR in the following text. Although it didn't surpass supervised learning methods, it offered a new approach.

Knowledge hypergraphs expand the concept of hyperedge for large-scale complex knowledge modeling and reasoning, showing potential in semantic understanding, intelligent QA, and multi-modal fusion. PLMs excel in knowledge extraction tasks, making them ideal for constructing knowledge hypergraphs due to their language understanding and multimodal processing capabilities.

Currently, several research endeavors have begun to explore the application of knowledge hypergraphs within RAG systems. In HyperGraphRAG \cite{ref20}, hypergraphs are built by extracting N-ary relations from text to accurately represent complex entity connections. LLMs handle end-to-end extraction, and the hypergraphs are stored in a bipartite graph database for efficient retrieval. Hyper-RAG \cite{ref21} uses PLMs to extract entities and relations, mapping them to the hypergraph. Hyperedge construction combines entity relationships with external evidence to verify reliability.

\begin{figure}
\includegraphics[width=\textwidth]{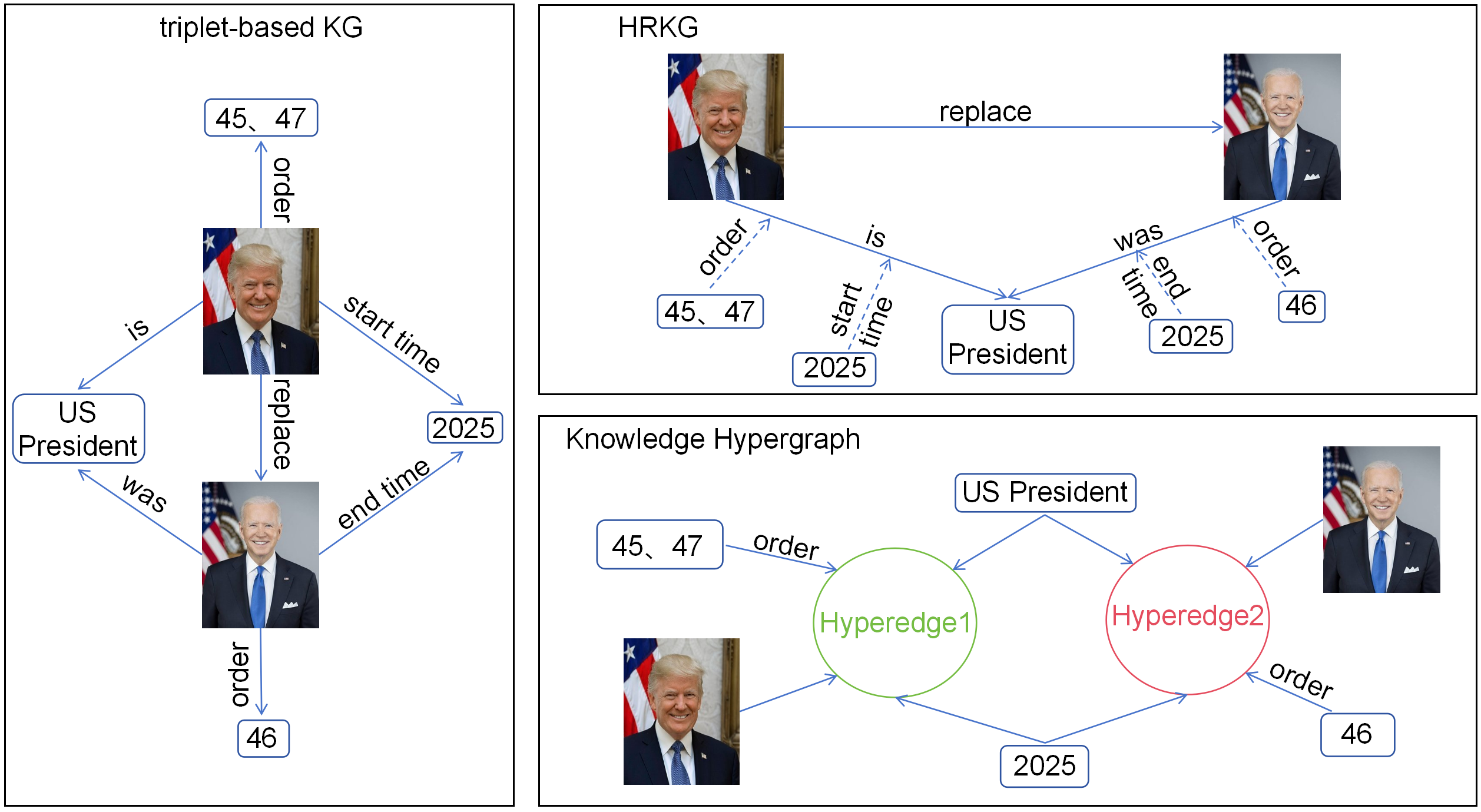}
\caption{Comparison of triplet-based KG, HRKG and Knowledge Hypergraph.} \label{fig2}
\end{figure}

Figure \ref{fig2} shows that triplet-based KG, HRKG and knowledge hypergraphs each have unique strengths and uses. Simple triplet-based KG suit binary relations, while HRKG, with added info, handle complex relations and queries. Knowledge hypergraphs are highly flexible but less mature. Choosing the right format is key to building efficient KG.

\section{Comparison of our solution with prior LLM-based HRKG constructions.}
There is limited work on constructing HRKG. Current LLM-based methods have insufficient accuracy and simple schemes. Therefore, we collaborate with multiple lightweight open-source LLMs and automatically optimize prompt words for specific tasks to extract knowledge from text and construct HRKG.

\begin{figure}
\includegraphics[width=\textwidth]{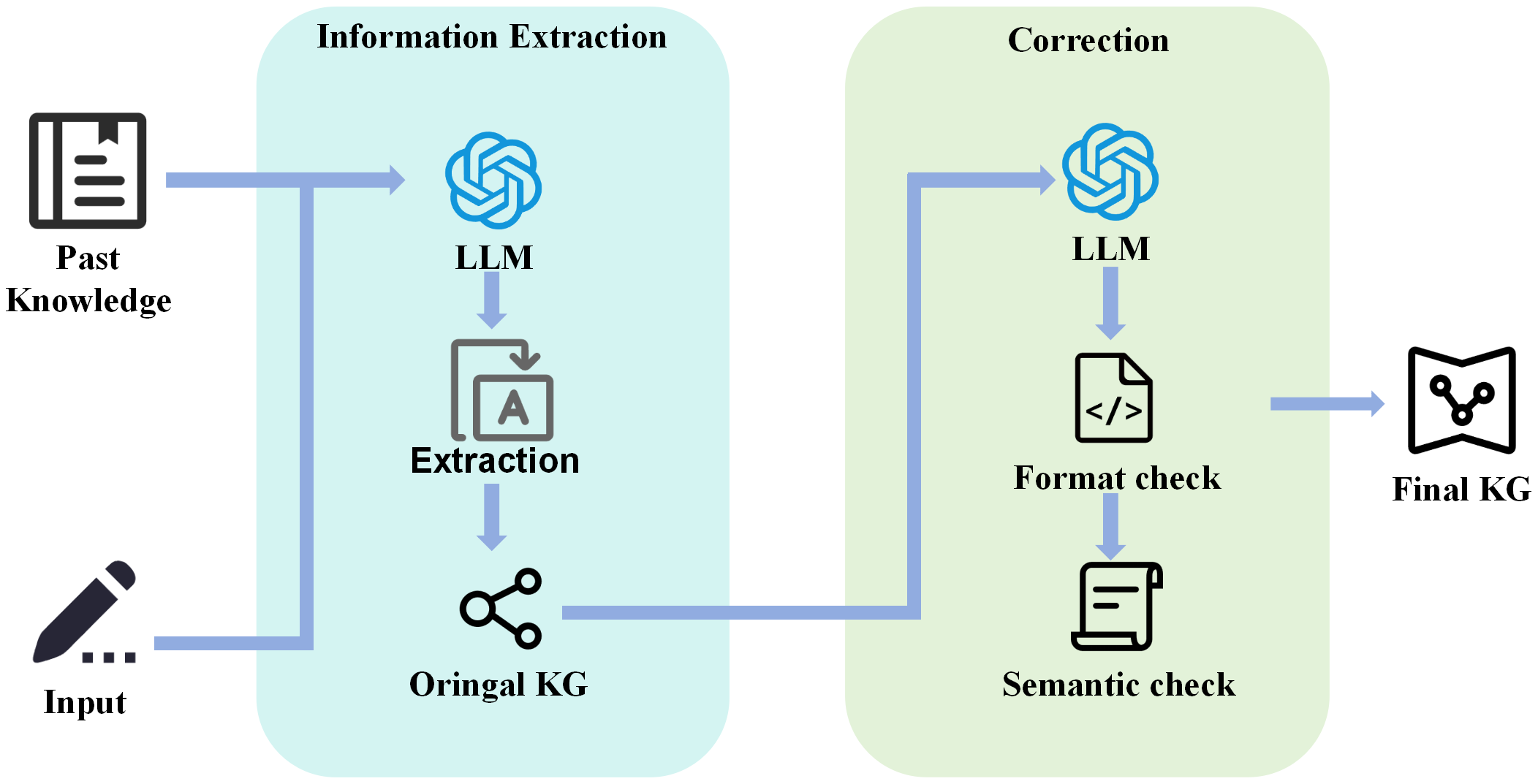}
\caption{The overall framework diagram of our solution is divided into prompt optimization, information extraction and correction modules. } \label{fig3}
\end{figure}

\subsection{Overall Framework}
The overall framework of this project is shown in Fig. \ref{fig3}, mainly including two parts: The hyper-relation extraction module extracts HRKG from the actual input by referring to the existing structured knowledge, and the HRKG correction module corrects the semantics and format of the HRKG extracted in the previous step based on the actual input text.
\subsection{Data set and Evaluation index}
HyperRED is a dataset used for Hyper-Relational Extraction, aiming to extract relational facts containing qualified information from text.
BERTScore \cite{ref22} is a text similarity evaluation tool based on the BERT pre-trained model. It calculates the context embedding representations of candidate texts and reference texts in the BERT model and uses cosine similarity to measure the semantic similarity between the two.
\subsection{Specific implementation details}
We select the LLama3.1:8B model for hyper-relation extraction, and the Qwen2.5:7B model for hyper-relation correction.
\subsection{Experimental result}
\begin{table}
\caption{The result of two schemes on BERTScore evaluations.}\label{tab2}
\centering
\begin{tabular}{|c|c|c|c|c|}
\hline
Framework & Model & Precision & Recall & F1\\
\hline
GCLR \cite{ref19} & GPT3.5 & 0.53 & 0.56 & 0.53\\
LLHKG(Ours) & LLama3.1:8B\&Qwen2.5:7B & 0.52 & 0.56 & 0.53\\
\hline
\end{tabular}
\end{table}
As shown in Table \ref{tab2}, under the BERTScore standard, our scheme is only 0.01 lower than the GCLR using GPT3.5. Our framework can improve the ability of lightweight LLMs to build HRKG to a level comparable to GPT3.5.

\section{Discussion}
PLMs have evolved from simple architectures to large-scale, high-performance models like the GPT and LLama series. Through unsupervised learning on massive text data, they have gained strong language understanding and generation capabilities, significantly advancing KG construction by efficiently extracting entities and relations from text and enriching KGs through generation and reasoning. However, current methods mainly rely on prompt engineering and fine-tuning. Fine-tuning requires high-quality data and significant computing power. Prompt engineering can mislead LLMs to generate inaccurate results if the prompts are biased or incomplete.

\section{Conclusion}
As a key branch of knowledge engineering, KG has become a core driver and important field for AI development. In this paper, we propose a new HRKG construction framework, LLHKG, which makes the HRKG construction capability of lightweight LLM comparable to that of GPT3.5.

In addition, this paper reviews KG construction technology, discusses integrating KGs with PLM, and highlights research challenges and potential solutions. Current progress shows that using PLM for KG construction is a mainstream approach with good results. However, both KG construction and LLMs face flaws and challenges. Future research directions in KG will likely include open-domain knowledge extraction using PLM, multi-source knowledge fusion, dynamic knowledge reasoning, and leveraging KG to address some limitations of LLMs.

\subsubsection{Acknowledgements} This work is funded in part by the Major Science and Technology Projects in Yunnan Province under Grant 202302AG050009 and 202302AB080014, in part by the Yunnan Fundamental Research Projects under Grant 202401AW070019, 202301AV070003.

%
%
%
%

\end{document}